\documentclass[pdflatex,sn-mathphys-num]{sn-jnl}

\usepackage{graphicx}%
\usepackage{multirow}%
\usepackage{amsmath,amssymb,amsfonts}%
\usepackage{amsthm}%
\usepackage{mathrsfs}%
\usepackage[title]{appendix}%
\usepackage{xcolor}%
\usepackage{textcomp}%
\usepackage{manyfoot}%
\usepackage{booktabs}%
\usepackage{algorithm}%
\usepackage{algorithmicx}%
\usepackage{algpseudocode}%
\usepackage{listings}%

\usepackage{graphics} 
\usepackage{epsfig} 
\usepackage{mathptmx} 
\usepackage{times} 

\usepackage{verbatim}
\usepackage{color}
\usepackage{soul}
\usepackage{wrapfig}
\usepackage[labelformat=simple]{subcaption}
\usepackage{balance}

\usepackage{multirow}
\usepackage{makecell}
\usepackage{hyperref}
\hypersetup{
    colorlinks=true,
    linkcolor=blue,
    filecolor=magenta,      
    urlcolor=cyan,
    pdftitle={Overleaf Example},
    pdfpagemode=FullScreen,
    }

\begin{document}

\title[Article Title]{Modeling and LQR Control of Insect Sized Flapping Wing Robot}


\author*[1]{\fnm{Daksh} \sur{Dhingra}}\email{dd292@uw.edu}

\author[2]{\fnm{Kadierdan} \sur{Kaheman}}

\author[1]{\fnm{Sawyer B.} \sur{Fuller}}

\affil*[1]{\orgdiv{ Department of Mechanical Engineering}, \orgname{University of Washington}, \orgaddress{ \city{Seattle},\state{WA}, \country{USA}}}

\affil[2]{\orgname{Dolby Laboratories (The work presented here is independent from author's work at Dolby Laboratories)}, \orgaddress{\city{San Francisco}, \state{CA}, \country{USA}}}


\abstract{Flying insects can perform rapid, sophisticated maneuvers like backflips, sharp banked turns,  and in-flight collision recovery. To emulate these in aerial robots weighing less than a gram, known as flying insect robots (FIRs), a fast and responsive control system is essential. To date, these have largely been, at their core, elaborations of proportional-integral-derivative (PID)-type feedback control. Without exception, their gains have been painstakingly tuned by hand. Aggressive maneuvers have further required task-specific tuning. Optimal control has the potential to mitigate these issues, but has to date only been demonstrated using approxiate models and receding horizon controllers (RHC) that are too computationally demanding to be carried out onboard the robot. Here we used a more accurate stroke-averaged model of forces and torques to implement the first demonstration of optimal control on an FIR that is computationally efficient enough to be performed by a microprocessor carried onboard. We took force and torque measurements from a 150 mg FIR, the UW Robofly, using a custom-built sensitive force-torque sensor, and validated them using motion capture data in free flight. We demonstrated stable hovering (RMS error of about 4 cm) and trajectory tracking maneuvers at translational velocities up to 25 cm/s using an optimal linear quadratic regulator (LQR)\footnotemark. These results were enabled by a more accurate model and lay the foundation for future work that uses our improved model and optimal controller in conjunction with recent advances in low-power receding horizon control to perform accurate aggressive maneuvers without iterative, task-specific tuning. 

\vspace{1.9cm}
\footnotetext{The video of the results can be accessed using: \url{www.youtube.com/watch?v=0o7j1nS2KHA}}
}




\maketitle
\section{Introduction}
\label{sec1}
Research in flapping wing insect-sized robots (FIRs) is motivated by their potential applications. These robots are small in size and are inexpensive to manufacture at a large scale, which makes them suitable for applications like detecting gas leaks, looking for survivors in disaster-prone areas, automated farm monitoring, running inspections on manufacturing lines, and weather monitoring. While still tethered and limited to operation inside a lab environment recent advances in tiny sensors and microcontrollers have brought them one step closer to achieving power~\cite{james2018liftoff} and sensor~\cite{Yash2022} autonomy.

Controlling FIRs presents significant challenges due to their highly nonlinear dynamics, manufacturing inconsistencies resulting in variability between robots, rapid wear and tear, and a high torque to moment of inertia ratio, approximately $10^3$ rad/sec$^2$, which leads to extremely fast dynamics.

Current state-of-the-art controllers for FIRs primarily utilize adaptive PID flight control systems for hovering~\cite{chirarattananon2014adaptive}. Despite their widespread use, these controllers require substantial ad-hoc tuning and are task-specific, often failing to consider actuator, state, and environmental constraints. Maneuvering beyond basic linear responses, such as perching~\cite{perchingwood2016} and somersault~\cite{KevinChenFlip} , requires a sliding mode controller combined with iterative learning of trajectory parameters. However, the parameters that were derived to perform an aggressive perch in~\cite{perchingwood2016} are specific to that task and cannot be applied to any other task.

Recent research has introduced optimal control strategies like modular Model Predictive Control (MPC)~\cite{2022AvikModularMPC}, which combines high-level MPC with a low-level controller for torque management, enabling operation beyond hovering. However, these systems have not been demonstrated on actual hardware for maneuvers beyond hovering. Additionally, data-driven MPC approaches like Tube-MPC~\cite{tagliabue2022robust} show promise for optimizing under actuator constraints and trajectory tracking for complex maneuvers such as ramps and infinity loops but remain too computationally intensive for implementation on sub-150 mg robots. Microprocessors small enough to be carried onboard, such as the 10 mg, 120 MHz STM32F4 used in the first wireless liftoff of an FIR, the UW Robofly in~\cite{james2018liftoff}, are capable of floating-point math operations. Nevertheless, their performance will be limited to a fraction of desktop capabilities, just a few hundred MHz, for the foreseeable future.

\begin{figure}[t]
    \centering
    \includegraphics[width=0.8\textwidth]{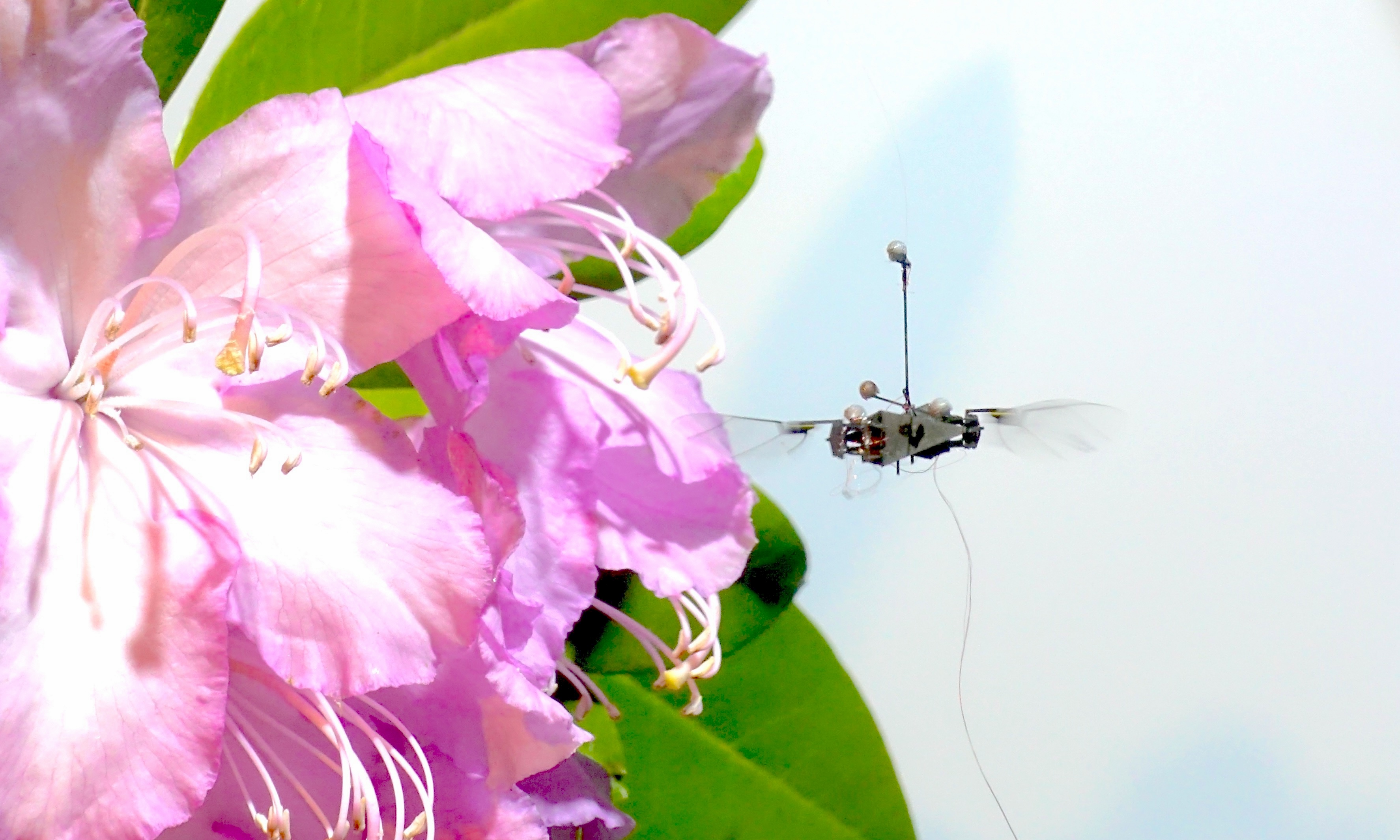}
    \caption{RoboFly, an insect-sized flapping robot weighing 146 milligrams, hovers next to a flower using feedback from motion capture cameras. The robot performs this hovering maneuver using the LQR controller reported in this work.}
    \label{Robofly_fig}
\end{figure}
\begin{figure*}[t]
    \centering
    \includegraphics[width=\textwidth]{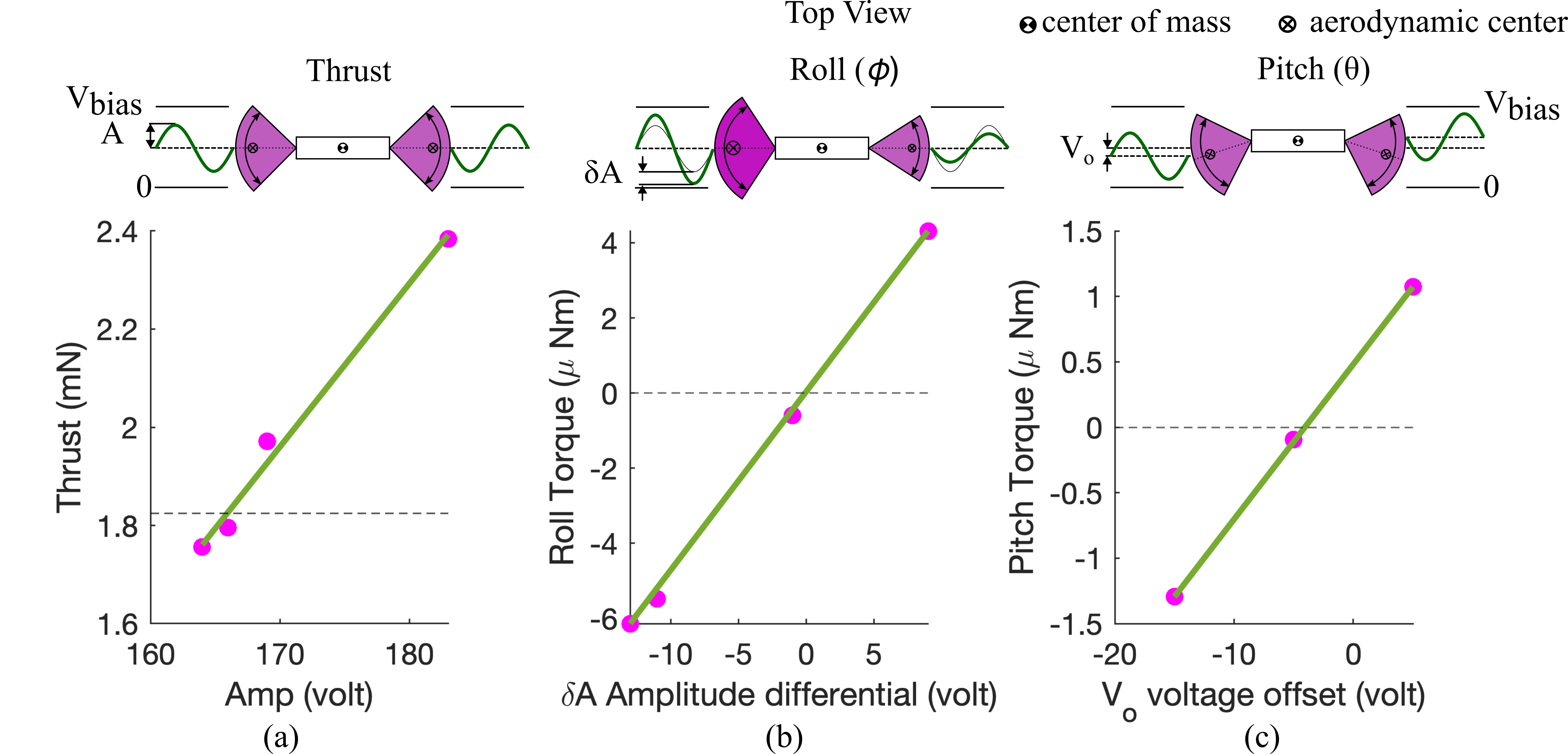}
    \caption{Torque and Thrust Generation Mechanism in FIRs: (Top) Inspired by the work in~\cite{ddhingraTrimming}, this figure shows that changing the signal parameters $\delta A$ and $V_o$ introduces roll and pitch torques, respectively. Here, $V_{bias}$ is the bias voltage. (Bottom) Mapping of (a) thrust, (b) roll torque, and (c) pitch torque of the RoboFly used in this work. The thrust mapping is obtained using a high-precision scale, while the torque mappings are obtained using the torque measurement device introduced in~\cite{ddhingraTrimming}. Pink dots represent the collected data points, and the green line represents the linear fit of the data. The corresponding equations for these linear fits are provided in Table~\ref{MAtable}.}
    \label{mapping_fig}
\end{figure*}
The primary focus of the work in~\cite{Bee++2023} is the precise tracking of the yaw angle, originally introduced in~\cite{chukewad2018new}. The PID controller used in~\cite{Bee++2023} is tuned based on the robot's dynamics. This enables the robot to hover and follow trajectories, such as an infinity loop. Our approach to controlling FIRs is based on the premise that an accurate model eliminates the need for laborious and unsatisfactory hand-tuning of PID gains. By framing the control problem within an optimal framework, we can design performance to maximize metrics like power efficiency or completion time. Using this model, we compute optimal gains around a fixed point for control. This was first demonstrated in~\cite{quad_LQR_2004}, where the performance of PID and LQR controllers were compared on quadrotors with the target to stabilizing the orientation angles. The LQR controller demonstrated faster response in reaching the reference signal from high initial angles. The recent development of low-cost and accurate torque measurement device~\cite{ddhingraTrimming} has improved our ability to map the torque characteristics of these robots accurately. Furthermore, innovations such as pre-stacked actuators~\cite{Prestack_Jafferis_2015} and standardized manufacturing processes~\cite{ChukewadTRO} have reduced variability in actuator performance and robot construction. In this work, we have made the following contributions:
\begin{enumerate}
    \item For the first time, we developed and validated a stroke-averaged first-principle model by comparing it with high-speed trajectories collected from a sub-150 mg robot.
    \item To the best of our knowledge, we present the first LQR implementation of controlled flight on an FIR. 
\end{enumerate}
This control strategy is computationally efficient, requiring a relatively small number of multiply-accumulate operations and trigonometric calculations per control step making it feasible for integration on tiny microcontrollers like the STM32F4, suitable for sub-150 mg robots.
We expect our model and LQR controller will be able to serve as integral elements in a fast onboard receding horizon optimal controllers, such as those discussed in~\cite{alavilli2024tinympc} and~\cite{englert2018software}, that can optimize under actuator limits and state constraints. With these receding horizon controllers, the presented model can enable more aggressive maneuvers due to the robots' high torque-to-inertia ratios, by scheduling the control gains for high translation speeds and attitude angles.

\section{Results}\label{sec2}

\begin{figure*}[t]
    \centering
    \begin{subfigure}[t]{0.38\textwidth}
    \centering
    \includegraphics[width=\textwidth]{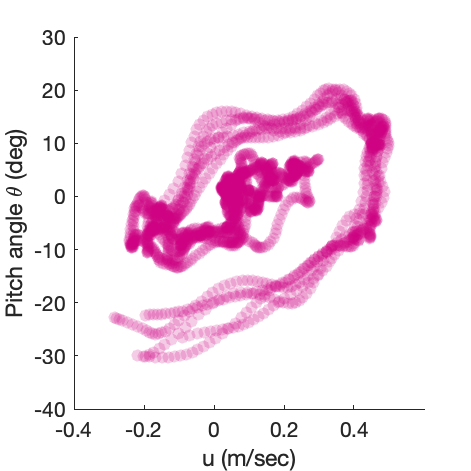}
     \end{subfigure}
     \begin{subfigure}[t]{0.38\textwidth}
     \centering
    \includegraphics[width=\textwidth]{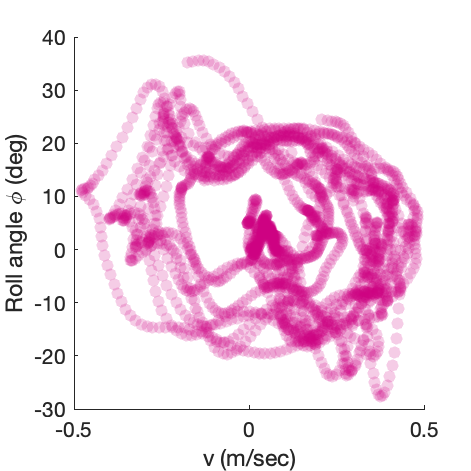}
     \end{subfigure}
    \caption{Visualization of the collected data: The graph shows the robot achieving high attitude angles greater than 30$^\circ$ and corresponding lateral/longitudinal speeds exceeding 0.4~m/s in the collected data. This highlights significant perturbations, which will be used to validate the stroke-averaged dynamics developed in this work. The color intensity on the graph represents the density of data points.}
    \label{fig_flightEnvelop}
\end{figure*}
\subsection{RoboFly}
RoboFly (shown in Fig.~\ref{Robofly_fig}) is a flapping-wing robot that weighs 150 mg. The robot features two piezoelectric actuators as muscles to flap its wings. These actuators are linked to a transmission mechanism that amplifies the actuators' displacement of approximately 200 $\mu$m to a wing motion of about 60$^\circ$. RoboFly has the capability to carry a payload up to 1.5 times its own weight and is powered via a wire-tether, which comprises four wires transmitting signals to operate the actuators.

RoboFly, like other piezo-actuated flapping wing robots~\cite{ma2013controlled}\cite{beeplus}, is operated by low-power 180~volt sinusoidal signals. It can generate roll torque, pitch torque, and thrust almost independently~\cite{AaronMapping2024}. The analog voltage signal is generated using the equation, 
\begin{align}
    V_{signal} = \frac{A + \delta A}{2}\sin \omega t + \frac{V_{bias}}{2} + \frac{V_o}{2}
\end{align} 
Here, $V_{bias}$ is a constant bias signal voltage of 250~volt supplied to the top layer of the actuator. As shown in Fig.~\ref{mapping_fig} (a), increasing the amplitude ($A$) of the sinusoidal signal increases the wing flapping amplitude, thereby generating greater thrust. Creating an amplitude differential ($\delta A$) between the wings, shown in Fig.~\ref{mapping_fig} (b), increases thrust on one side while decreasing it on the other, which produces roll torque. Pitch torque (Fig.~\ref{mapping_fig} (c)) is generated by adjusting the wing flapping either forward or backward relative to the robot's body through a voltage offset ($V_o$) applied to the sinusoidal signal.
\subsection{Theoretical Model}

The dynamics of a flapping wing robot of the size of RoboFly can be defined by the same first principle model as a quadrotor~\cite{Bee++2023}. Using the convention of $(X,Y,Z)$ as coordinates in an inertial frame and $(x_b, y_b, z_b)$ as coordinates in the body frame, first-order differential equations of the body velocity $ V_b = [u, v, w]^T$ of the robot is defined by 
\begin{align}
 \label{xcoordinate_eq}
 \dot{u} =&  g \sin\theta + f_{a_1}  - (qw - rv)\\
   \label{ycoordinate_eq}
 \dot{v} =& - g \cos\theta \sin\phi + f_{a_2}  - (rv - pw)\\
 \dot{w} =& - g \cos\theta \cos\phi + f_{a_3}  - (pv - qu) + \frac{\Gamma}{m+m_M}
 \label{zcoordinate_eq}
\end{align}
 Here, $ \omega_b = [p, q, r]^T$ is the body angular velocity, $g$ is the acceleration due to gravity, $m$ is the mass of the robot and, $m_M$ is the mass of the MoCap markers. Rotation is defined by the widely used 321 rotation matrix. The 321 rotation follows the yaw ($\psi$) $\rightarrow$ pitch ($\theta$) $\rightarrow$ roll ($\phi$) sequence. $\Gamma$ is the thrust applied to the robot in the positive $z_b$ direction. $ F_a = [f_{a_1}, f_{a_2}, f_{a_3}]^T$ is the unmodeled dynamic force, which includes stroke averaged aerodynamic force in body frame caused by the drag due to flapping wings. Similarly, rotational dynamics can be defined by the first-order differential equation in body angular velocity ($ \omega_b =[p, q, r]^T$).
\begin{align}
\dot{p} =& L + \frac{\tau_r}{J_{xx}} - \frac{J_{zz}-J_{yy}}{J_{xx}} qr\\
\dot{q} =& M + \frac{\tau_p}{J_{yy}} - \frac{J_{xx}-J_{zz}}{J_{yy}} rp\\
\dot{r} =& N - \frac{J_{yy}-J_{xx}}{J_{zz}} pq
\label{rcoordinate_eq}
\end{align}

Here, $  J = diag([J_{xx}, J_{yy}, J_{zz}])$ is the diagonal moment of inertia matrix of the robot. $ \tau_c = [\tau_r, \tau_p, 0]^T$ is defined as the input torque vector comprising of roll torque ($\tau_r$) about $x_b$ and pitch torque ($\tau_p$) about $y_b$. $ \tau_a = [L, M, N]^T$ is the unmodeled dynamic moment, which also includes averaged aerodynamic torque caused by the drag due to flapping wings.
\subsubsection{Model Parameters} 
We precisely measured the robot's parameters using a sensitive, custom-build force torque sensor. Results are tabulated in Table~\ref{MAtable}. The mass of the robot was determined using a high-precision balance with a resolution of 0$.$1~mg. To estimate the robot's moment of inertia, we measured the mass of its various components, including the piezoelectric actuators, wings, airframes, and motion capture markers. These measurements were then input into a CAD model to calculate the robot's moment of inertia matrix. The high-precision scale also facilitated in the calculation of thrust mapping, where the thrust generated by the robot at specific flapping amplitudes ($A$) was recorded. During the experiment, the wings were flapped for a period of 1~sec at the constant frequency of 180~hz. The scale can take accurate stroke averaged measurements of thrust. The test setup also makes sure that the robot is away from any surrounding objects to avoid ground effects. We employed a least squares fit method to model the thrust based on this data (Fig.~\ref{mapping_fig}-left). The learned linear fit of the thrust mapping is shown in Table~\ref{MAtable}. Given that the lifetime of these robots is about 10 minutes~\cite{malka2014}, we aimed to minimize the total operating time to prevent mechanical fatigue. Therefore, for the mapping of thrust and torques, we took only two or three measurements to establish a trendline that can be used in the model.

To measure the torque response to different voltage inputs, that is, torque mapping, we used a device similar to the one in~\cite{AaronMapping2024}. By applying inputs $V_o$  and $\delta V$, we generate roll and pitch torques, respectively. These torques induced angular deflections on the device, which are linearly correlated with the applied torques, which was observed in practice in~\cite{AaronMapping2024}. These deflections were accurately measured using the motion capture system, allowing us to map torques effectively (Fig.~\ref{mapping_fig}-middle and right). This comprehensive approach to parameter measurement ensures a robust foundation for our model.

\subsection{Trajectory Data}

We collected 8 seconds of trajectory data from 7 separate flights with wings flapping at a frequency of 180 Hz, controlled by a PID flight controller~\cite{ChukewadTRO}. To capture flight perturbations, we set the desired points away from the robot's initial position, focusing on collecting more data with perturbations in lateral, longitudinal, and vertical dynamics. To avoid capturing too much stable hovering data, most flight trajectories were shortened to the duration required for the robot to reach the set positions. The robot was equipped with four retro-reflective markers to track its position and orientation through a motion capture system comprised of four Prime 13 cameras by OptiTrak, Inc., Salem, OR. Position and quaternions from the motion capture system running at 240 hz were used to calculate $ V_b$, $ \omega_b$, and Euler angles of the robot offline.

\subsubsection{Body Offset}
\label{body-offset_sec}
A critical difference between the robot's trajectory data and the modeled dynamics from equations~\ref{xcoordinate_eq}-\ref{rcoordinate_eq} can stem from the misalignment between the body z-axis, as defined in the motion capture software, and the robot's thrust vector. The dynamics, detailed in equations~\ref{xcoordinate_eq}-\ref{zcoordinate_eq}, assume that the thrust vector is perfectly aligned with the body z-axis, an assumption that may not hold in practice. This misalignment issue occurs because the thrust vector's direction is not known at the time that the robot's body coordinates are defined in the motion capture software. A tilted thrust vector introduces lateral and longitudinal forces. To reduce this error, we redefined the body coordinate system after performing a short 0.3~s uncontrolled trimmed flight. Since during a trimmed flight, the robot takes off approximately vertically, so its trajectory can be used to estimate the direction of the thrust vector and therefore align the z-axis of the body coordinate system to that direction. This, however, is based on the assumption that the body coordinate system is not redefined in between the experiments and remains the same throughout the data collection and control process.  
\begin{table}[b]
\centering
\begin{tabular}[h!]{|c|c|c|} 
 \hline
\textbf{Parameter} & \textbf{Symbol} &\textbf{Value}\\
  \hline
Mass of the robot & $m$ & $150\times 10^{-6}$~kg\\
\hline
Mass of the MoCap markers & $m_M$ & $36\times 10^{-6}$~kg\\
  \hline
Moment of inertia & $J$ & diag([$3.12\times 10^{-9}$, $2.97\times 10^{-9}$, $0.55\times 10^{-9}$])~kg.m$^2$ \\
  \hline
Thrust & $\Gamma$ & $3.27\times 10^{-5} A-0.0024$~N\\
  \hline
Roll Torque & $\tau_r$ & $0.48\times 10^{-6} \delta A$~Nm\\
  \hline
Pitch Torque & $\tau_p$ & $0.11\times 10^{-6} V_o$~Nm\\
\hline
\end{tabular}
\caption{Measured parameters of RoboFly. Calculated Moment of inertia $J$ is a diagonal matrix of moment of inertia about the principal axes. Mapping from $A$ to $\Gamma$, $\delta A$ to $\tau_r$, and $V_o$ to $\tau_p$ are the equations of linear fit from Fig~\ref{mapping_fig} (a),(b) and (c) respectively.  }
\label{MAtable}
\end{table}

\subsubsection{Visualization of Collected Data}
Unlike traditional aircraft, where the flight envelope is defined by changes in velocity against variations in angle of attack, flapping wing robots exhibit continuous angle of attack variations throughout the flapping cycle. Therefore, we determined that representing the flight state in terms of angular orientation and velocity would be more pertinent characterization. Characterizing in this way  elucidates the robot's ability to maintain controlled flight across different tilt angles and the associated longitudinal/lateral speeds at these angles, which are critical for maneuverability. A robot that can sustain higher tilt angles and translational speeds in controlled flight is indicative of superior maneuverability, enabling it to execute tighter turns.  As shown in Fig.~\ref{fig_flightEnvelop}, based on data collected from the flight trajectories of the RoboFly, the robot is able to get to attitude angles of approximately 30$^{\circ}$ and body velocity of around 0$.$4~m/sec.

\subsection{Controller Implementation}

\begin{figure}[t]
    \centering
    \includegraphics[width=\textwidth]{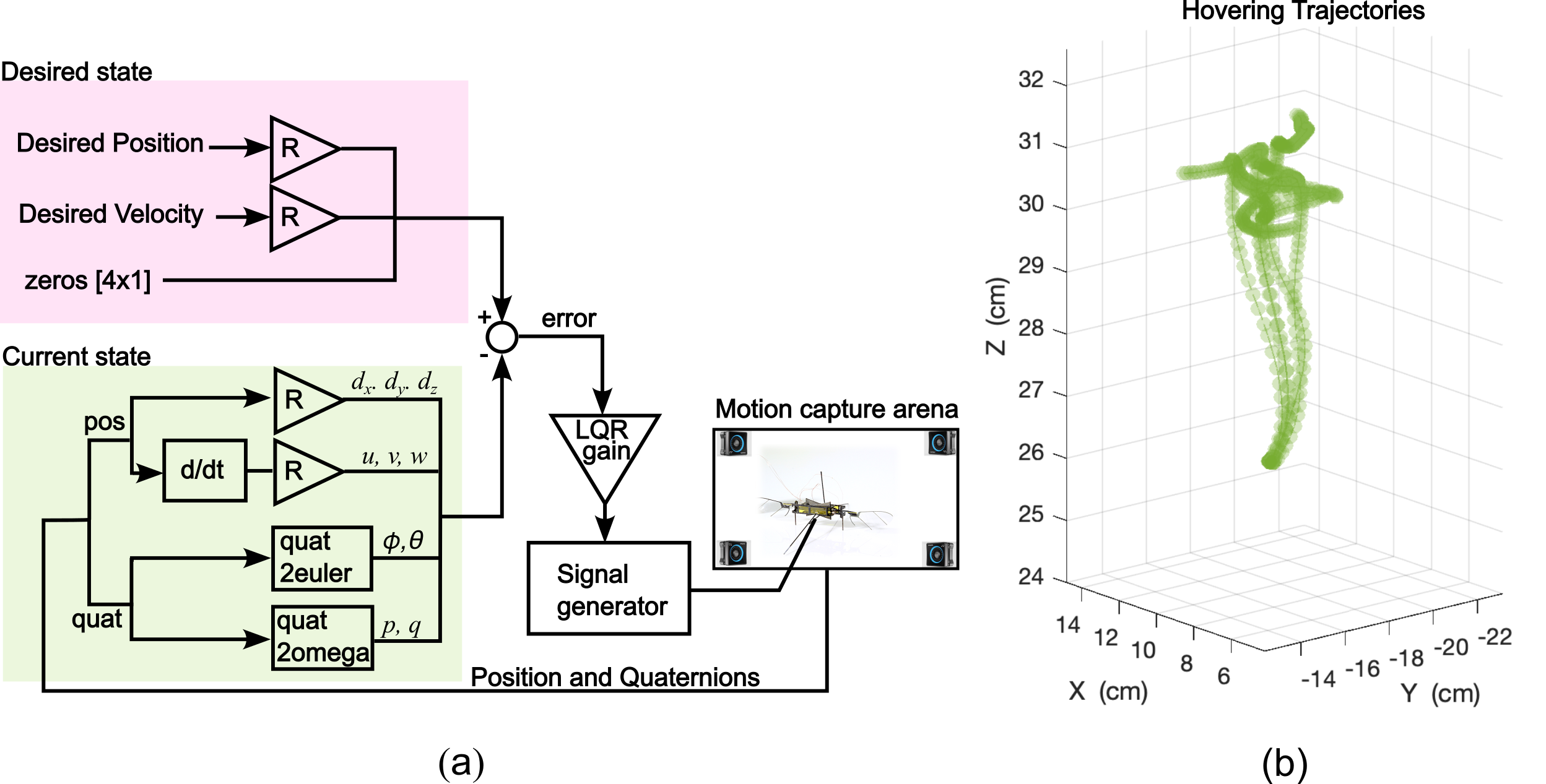}
    \caption{LQR control loop and hovering trajectories: (a) The LQR control loop used to perform hovering and trajectory tracking maneuvers. Here, R represents the 3-2-1 rotation matrix. (b) The plot displays five different hovering trajectories, showing the robot maintaining a stable attitude and remaining close to the starting position. The mean RMS error and standard deviation for the trajectories are 4.17$\pm$ 0.37~cm.  }
    \label{control_figure}
\end{figure}

In our experiments, we did not control the yaw rotation of the robot. Hence the LQR was designed to optimize the dynamics in the body coordinate system, which are independent of the yaw rotation. The state vector of the robot is defined by $ \sigma = [d_x, d_y, d_z, u, v, w, \phi, \theta, p, q]^T$. Here,  $ d = [d_x, d_y, d_z]^T$ is the position in body coordinates and $ V_b = \dot{d} = [u, v, w]$. The controller assumes that the angular velocity about $z_b$, denoted as $r$, which arises from manufacturing uncertainties, remains constant. Its value affects fictitious forces and torques in body coordinates that appear in equations~\ref{xcoordinate_eq}-\ref{rcoordinate_eq}. The controller calculates the inputs $u^* = [A,\delta A,  V_o]$, which directly controls the acceleration in $z_b$, torque about $x_b$, and torque about $y_b$ axis.
The $Q$ matrix used in our experiments is a diagonal matrix, $Q = diag([0.02, 0.02, 0.01, 0.1, 0.1, 0.1,  1, 1,  4, 4])$; $R$ is also a diagonal matrix, $R = diag([2,1,1])$. The ratio of $Q$ and $R$ matrix used here was obtained with the knowledge of our model and only one experimental flight. 
The feedback loop includes a motion capture system that provides state feedback in terms of the robot's position and orientation (expressed as quaternions). This data feeds into a Simulink real-time system in which the body angular velocities, euler angles, and velocity in world frame are calculated. The controller receives the desired position in the world coordinates. Finally, the resulting error in the world coordinates is converted to the body coordinates and is multiplied by the pre-calculated LQR gain matrix $k$ to determine the control inputs, $u =-k (\sigma_{des}-\sigma)$, for the robot. Control loop used for our experiments is shown in Fig.~\ref{control_figure}(a).

\begin{figure}[t]
    \centering
    \includegraphics[width=\textwidth]{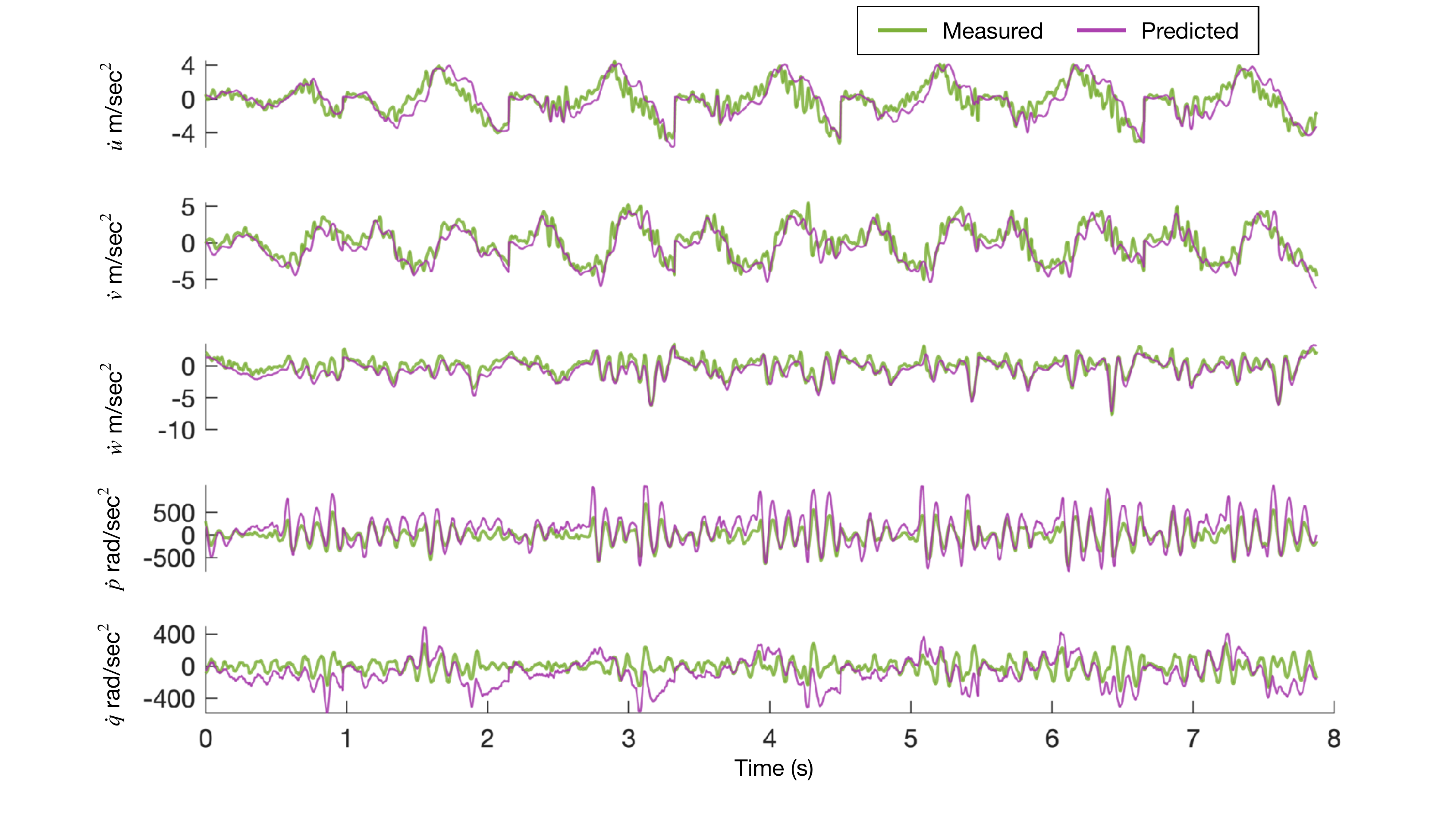}
    \caption{Model validation plots: Measured accelerations (green) from the RoboFly trajectories plotted with the predicted accelerations (pink) calculated using the theoretical model.}
    \label{Model_figure}
\end{figure}

\begin{figure*}
    \centering
    \includegraphics[width=\textwidth]{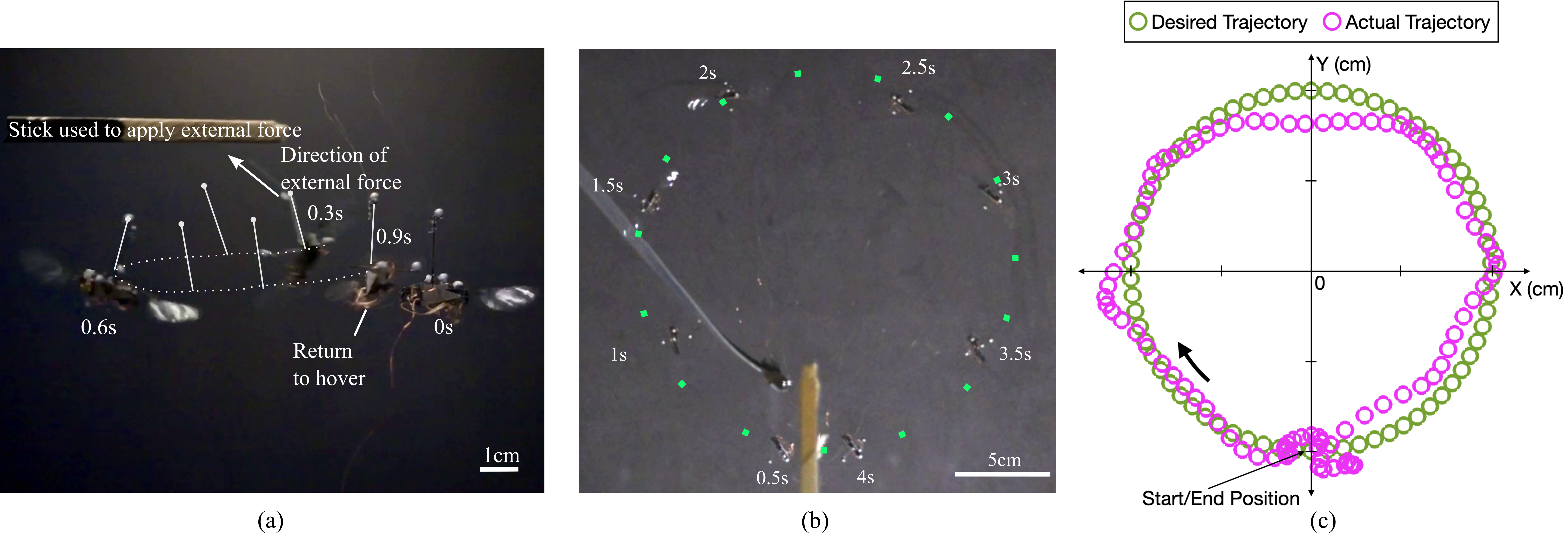}
\caption{ Results of LQR implementation on the RoboFly: (a) Response to the external disturbance applied via kevlar thread is shown in the photo composite. The robot does a recovery maneuver to get back to the stable attitude and ultimately flies to the desired position.(b) The photo composite shows the RoboFly tracking a 10~cm radius circular trajectory over a 4.5-second flight using a constant pre-calculated LQR gain. (c) The desired trajectory, depicted in green, was provided to the controller in the form of position and velocity set points while magenta is the actual trajectory followed by the robot. The RMS error for this maneuver for x-y position tracking is 2.8 cm. The video link of the experiments is shown in the abstract page.}
    \label{traj_track_figure}
\end{figure*}
\subsection{Model Validation}
For simplicity, the model we used in this work excludes damping and drag components, thus setting the force and moment vectors $[f_{a_1}, f_{a_2}, f_{a_3}]^T$ and $[L, M, N]^T$ to zero in equations (\ref{xcoordinate_eq}-\ref{rcoordinate_eq}). We validated the theoretical model accelerations by comparing them with measured acceleration data from flight tests, as depicted in Fig.~\ref{Model_figure}. This comparison includes seven separate flight trajectories stacked in time.

The model can predict translational accelerations in the body coordinate system, represented by $[\dot{u}, \dot{v}, \dot{w}]^T$, with the root mean squared ($L^2$) error of 53.4 m/sec$^2$, 56.9 m/sec$^2$, and 36.7 m/sec$^2$, respectively. The errors in rotational accelerations, $[\dot{p}, \dot{q}]^T$, are 9.2e$^3$ rad/sec$^2$ and 7.6e$^3$ rad/sec$^2$, respectively. The more substantial errors in the rotational domain can largely be attributed to the aforementioned aspect of small flight vehicles that their angular accelerations are large.

We believe, and our results show, that the model is still adequate for controller design purposes. This is based on two key considerations: Firstly, the actuation delay for the rotational system is minimal, as rotational acceleration occurs almost instantaneously once torque is applied. Secondly, the robot will eventually have a gyroscope onboard which is capable of providing rapid rotational velocity feedback at 1 to 16 kHz, which will significantly enhance the ability to perform rapid feedback corrections. Thus, even with its limitations, this model provides a sufficient foundation for developing an effective controller.

\subsection{Hovering}

The initial task was to hover around a desired position. We did five such flights, each lasting 2 seconds, and the robot managed to hover using the pre-calculated LQR gain, with RMS errors of 4.2 cm, 3.8 cm, 4.05 cm, 4.03 cm, and 4.8 cm. For hovering tasks, a PID controller, as referenced in~\cite{ma2013controlled} and~\cite{fuller2019four}, outperforms this with an average RMS error of 2 cm. We think this is due to the fact that PID controllers are manually tuned for specific tasks, whereas the LQR controller used here was only tuned once for determining the Q and R matrices and is more general, as it can be linearized about different states. Five hover trajectories are shown in Fig.~\ref{control_figure}(b).

\subsection{Response to External Disturbance}

To prevent crashes during our experiments, we suspend the robot using a lightweight Kevlar thread. There is slack in the Kevlar thread in all our experiment videos, which demonstrates that the thread does not exert any force on the robot. However, in this particular experiment, we intentionally applied force by pulling the robot with the Kevlar thread to test the controller's response to external disturbances. This causes the accelerations $>$2$.$5g.  As show in Fig.~\ref{traj_track_figure}(a), the robot was able to stabilize itself and flies towards the desired position.

\subsection{Trajectory Tracking}
Here the robot was asked to follow a pre-computed circular trajectory of 10~cm radius. The robot was able to follow the given trajectory in a 4.5-second flight with an RMS error of 2.8~cm in x-y position tracking. Photo composite of the maneuver is shown in Fig.~\ref{traj_track_figure}(b). The desired waypoints on the trajectory were given in the form of position and velocity set points. The controller used the same Q and R matrices as the hovering maneuver. 
For comparison, in~\cite{tagliabue2022robust} authors tracked a 5~cm  radius circular trajectory with the reported x-y position error of 1.8~cm. Our greater position tracking error is likely due to a much higher flight velocity (25~cm$/$sec vs a maximum speed of 5.2 cm$/$s in~\cite{tagliabue2022robust}) and a larger circular trajectory, which would result in larger disturbances by the wire tether.

\section{Discussion}

In our study, we developed and validated a theoretical model of the UW RoboFly. Utilizing this model, we successfully implemented an infinite horizon Linear Quadratic Regulator (LQR) control strategy. This enabled us to achieve stable hovering, recovery maneuver, and trajectory tracking using the pre-calculated LQR gain. While its RMS position error was higher than other recent reports, it was following a much faster trajectory. Notably, our controller works with minimal computational demands, making it ideal for integration into microcontrollers suited for tiny robotic platforms. 

\subsection{Limitations and Future Work}
This work provides a foundation for some important next steps. First, the single LQR gain could be replaced with one that is ``gain-scheduled" for linearizations about different states, such as forward flight or flight into the wind. This would also impose a minimal computational load on any conceivable microcontroller. Second, collecting data over a broader flight envelope could improve model identification and therefore flight control. Third, our current controller does not perform any sort of adaptivity. Future models could account for the offset between the thrust vector and the body-z axis and estimate the drag model using the discrepancy between expected and actual translational velocity as input\cite{ma2013controlled,fuller2019four}. Such refinements are expected to yield even lower RMS errors for specific tasks and potentially enable more agile maneuvers.

While gain scheduling has the potential to broaden the envelope in which the LQR controller operates well, it does not take into account an important aspect of FIRs. This is that they are subject to constraints on the magnitude of outputs that the actuators can produce. A more advanced control technique, known as Receding Horizon Control (RHC) or Model Predictive Control (MPC), can accurately factor those into the optimization process. Although~\cite{tagliabue2022robust} applied this technique, their controllers were too demanding for a microcontroller. We expect that the advances that were used in Tiny-MPC~\cite{alavilli2024tinympc} to perform RHC on the 120 MHz microcontroller onboard the the 30~g Crazyflie helicopter could be adapted to the different dynamics and constraints of the Robofly. By taking into account actuator constraints and any improved system characterization, more precise and more aggressive agile maneuvers should be possible without hand-tuning.

\section{Methods}\label{sec11}
\subsection{Infinite Horizon LQR}
Infinite horizon LQR controller~\cite{andersonoptimal2007} optimizes the quadratic cost function subject to linearized dynamics constraints. In this work we use the continuous time formulation of LQR. 
\begin{align}
    J =& \int_0^{\infty} \sigma(t)^T  Q \sigma(t) + u(t)^T  R u(t)\\
    \text{subject to }  \dot{\sigma}(t) =& A_d \sigma(t) +   B_d u(t)
\end{align}
Here, $ \sigma(t)$ is the state vector of the robot at time $t$, $A_d$ and $B_d$ are the linearized dynamics matrices. If the system is completely stabilizable, we can write this optimal control problem in terms of the Hamiltonian function, which incorporates both the system dynamics and the cost function. The Hamiltonian for the LQR problem is given by $H(x,u,\lambda) = \sigma^TQ\sigma + u^TRu + \lambda^T(A_d\sigma+B_du)$. $\lambda$ is the costate vector, $Q$ and $R$ are positive semi-definite and positive definite matrices respectively. By taking the derivative of Hamiltonian with respect to $\sigma$, $u$, and $\lambda$ and setting it to zero we get the closed loop dynamics as,
\begin{align}
\begin{bmatrix}
\dot{\sigma} \\
\dot{\lambda}
\end{bmatrix}=
\begin{bmatrix}
    A & -BR^{-1}B^T\\
    -Q & -A^T\\
\end{bmatrix}
\begin{bmatrix}
    \sigma\\
    \lambda
\end{bmatrix}
\label{closed_loop_dynamics}
\end{align}
The steady state solution of the Riccati equation can be described in terms of the eigenvectors of the Hamiltonian matrix in the closed loop dynamic equation~\ref{closed_loop_dynamics} associated with the negative real part eigen values. These negative real part eigenvalues of the matrix are also the eigenvalues of the closed-loop matrix $A - BR ^{-1}BP$. The feedback gain $k = -R^{-1}B'P$ can be obtained using $P$ as the Schur form of the closed loop dynamics matrix such that: 
\begin{align}
u(t) =& -k(\sigma_{des}(t)- \sigma(t))
\end{align}
Here $\sigma_{des}(t)$ is the desired state of the robot at time t. 
\backmatter

\bmhead{Acknowledgements}
We would like to thank Y. M. Chukewad for insightful discussions about the Robofly.






\bibliography{main}


\begin{thebibliography}{21}
\ifx \bisbn   \undefined \def \bisbn  #1{ISBN #1}\fi
\ifx \binits  \undefined \def \binits#1{#1}\fi
\ifx \bauthor  \undefined \def \bauthor#1{#1}\fi
\ifx \batitle  \undefined \def \batitle#1{#1}\fi
\ifx \bjtitle  \undefined \def \bjtitle#1{#1}\fi
\ifx \bvolume  \undefined \def \bvolume#1{\textbf{#1}}\fi
\ifx \byear  \undefined \def \byear#1{#1}\fi
\ifx \bissue  \undefined \def \bissue#1{#1}\fi
\ifx \bfpage  \undefined \def \bfpage#1{#1}\fi
\ifx \blpage  \undefined \def \blpage #1{#1}\fi
\ifx \burl  \undefined \def \burl#1{\textsf{#1}}\fi
\ifx \doiurl  \undefined \def \doiurl#1{\url{https://doi.org/#1}}\fi
\ifx \betal  \undefined \def \betal{\textit{et al.}}\fi
\ifx \binstitute  \undefined \def \binstitute#1{#1}\fi
\ifx \binstitutionaled  \undefined \def \binstitutionaled#1{#1}\fi
\ifx \bctitle  \undefined \def \bctitle#1{#1}\fi
\ifx \beditor  \undefined \def \beditor#1{#1}\fi
\ifx \bpublisher  \undefined \def \bpublisher#1{#1}\fi
\ifx \bbtitle  \undefined \def \bbtitle#1{#1}\fi
\ifx \bedition  \undefined \def \bedition#1{#1}\fi
\ifx \bseriesno  \undefined \def \bseriesno#1{#1}\fi
\ifx \blocation  \undefined \def \blocation#1{#1}\fi
\ifx \bsertitle  \undefined \def \bsertitle#1{#1}\fi
\ifx \bsnm \undefined \def \bsnm#1{#1}\fi
\ifx \bsuffix \undefined \def \bsuffix#1{#1}\fi
\ifx \bparticle \undefined \def \bparticle#1{#1}\fi
\ifx \barticle \undefined \def \barticle#1{#1}\fi
\bibcommenthead
\ifx \bconfdate \undefined \def \bconfdate #1{#1}\fi
\ifx \botherref \undefined \def \botherref #1{#1}\fi
\ifx \url \undefined \def \url#1{\textsf{#1}}\fi
\ifx \bchapter \undefined \def \bchapter#1{#1}\fi
\ifx \bbook \undefined \def \bbook#1{#1}\fi
\ifx \bcomment \undefined \def \bcomment#1{#1}\fi
\ifx \oauthor \undefined \def \oauthor#1{#1}\fi
\ifx \citeauthoryear \undefined \def \citeauthoryear#1{#1}\fi
\ifx \endbibitem  \undefined \def \endbibitem {}\fi
\ifx \bconflocation  \undefined \def \bconflocation#1{#1}\fi
\ifx \arxivurl  \undefined \def \arxivurl#1{\textsf{#1}}\fi
\csname PreBibitemsHook\endcsname

\bibitem[\protect\citeauthoryear{James et~al.}{2018}]{james2018liftoff}
\begin{bchapter}
\bauthor{\bsnm{James}, \binits{J.}},
\bauthor{\bsnm{Iyer}, \binits{V.}},
\bauthor{\bsnm{Chukewad}, \binits{Y.}},
\bauthor{\bsnm{Gollakota}, \binits{S.}},
\bauthor{\bsnm{Fuller}, \binits{S.B.}}:
\bctitle{Liftoff of a 190 mg laser-powered aerial vehicle: The lightest wireless robot to fly}.
In: \bbtitle{2018 IEEE International Conference on Robotics and Automation (ICRA)},
pp. \bfpage{1}--\blpage{8}
(\byear{2018}).
\bcomment{IEEE}
\end{bchapter}
\endbibitem

\bibitem[\protect\citeauthoryear{Talwekar et~al.}{2022}]{Yash2022}
\begin{bchapter}
\bauthor{\bsnm{Talwekar}, \binits{Y.P.}},
\bauthor{\bsnm{Adie}, \binits{A.}},
\bauthor{\bsnm{Iyer}, \binits{V.}},
\bauthor{\bsnm{Fuller}, \binits{S.B.}}:
\bctitle{Towards sensor autonomy in sub-gram flying insect robots: A lightweight and power-efficient avionics system}.
In: \bbtitle{2022 International Conference on Robotics and Automation (ICRA)},
pp. \bfpage{9675}--\blpage{9681}
(\byear{2022}).
\doiurl{10.1109/ICRA46639.2022.9811918}
\end{bchapter}
\endbibitem

\bibitem[\protect\citeauthoryear{Chirarattananon et~al.}{2014}]{chirarattananon2014adaptive}
\begin{barticle}
\bauthor{\bsnm{Chirarattananon}, \binits{P.}},
\bauthor{\bsnm{Ma}, \binits{K.Y.}},
\bauthor{\bsnm{Wood}, \binits{R.J.}}:
\batitle{Adaptive control of a millimeter-scale flapping-wing robot}.
\bjtitle{Bioinspiration \& biomimetics}
\bvolume{9}(\bissue{2}),
\bfpage{025004}
(\byear{2014})
\end{barticle}
\endbibitem

\bibitem[\protect\citeauthoryear{Graule et~al.}{2016}]{perchingwood2016}
\begin{barticle}
\bauthor{\bsnm{Graule}, \binits{M.A.}},
\bauthor{\bsnm{Chirarattananon}, \binits{P.}},
\bauthor{\bsnm{Fuller}, \binits{S.B.}},
\bauthor{\bsnm{Jafferis}, \binits{N.T.}},
\bauthor{\bsnm{Ma}, \binits{K.Y.}},
\bauthor{\bsnm{Spenko}, \binits{M.}},
\bauthor{\bsnm{Kornbluh}, \binits{R.}},
\bauthor{\bsnm{Wood}, \binits{R.J.}}:
\batitle{Perching and takeoff of a robotic insect on overhangs using switchable electrostatic adhesion}.
\bjtitle{Science}
\bvolume{352}(\bissue{6288}),
\bfpage{978}--\blpage{982}
(\byear{2016})
\doiurl{10.1126/science.aaf1092}
{\href{https://arxiv.org/abs/https://www.science.org/doi/pdf/10.1126/science.aaf1092}{{https://www.science.org/doi/pdf/10.1126/science.aaf1092}}}
\end{barticle}
\endbibitem

\bibitem[\protect\citeauthoryear{Chen et~al.}{2021}]{KevinChenFlip}
\begin{barticle}
\bauthor{\bsnm{Chen}, \binits{Y.}},
\bauthor{\bsnm{Xu}, \binits{S.}},
\bauthor{\bsnm{Ren}, \binits{Z.}},
\bauthor{\bsnm{Chirarattananon}, \binits{P.}}:
\batitle{Collision resilient insect-scale soft-actuated aerial robots with high agility}.
\bjtitle{IEEE Transactions on Robotics}
\bvolume{37}(\bissue{5}),
\bfpage{1752}--\blpage{1764}
(\byear{2021})
\doiurl{10.1109/TRO.2021.3053647}
\end{barticle}
\endbibitem

\bibitem[\protect\citeauthoryear{De et~al.}{2022}]{2022AvikModularMPC}
\begin{barticle}
\bauthor{\bsnm{De}, \binits{A.}},
\bauthor{\bsnm{McGill}, \binits{R.}},
\bauthor{\bsnm{Wood}, \binits{R.J.}}:
\batitle{An efficient, modular controller for flapping flight composing model-based and model-free components}.
\bjtitle{The International Journal of Robotics Research}
\bvolume{41}(\bissue{4}),
\bfpage{441}--\blpage{457}
(\byear{2022})
\doiurl{10.1177/02783649211063225}
{\href{https://arxiv.org/abs/https://doi.org/10.1177/02783649211063225}{{https://doi.org/10.1177/02783649211063225}}}
\end{barticle}
\endbibitem

\bibitem[\protect\citeauthoryear{Tagliabue et~al.}{2022}]{tagliabue2022robust}
\begin{botherref}
\oauthor{\bsnm{Tagliabue}, \binits{A.}},
\oauthor{\bsnm{Hsiao}, \binits{Y.-H.}},
\oauthor{\bsnm{Fasel}, \binits{U.}},
\oauthor{\bsnm{Kutz}, \binits{J.N.}},
\oauthor{\bsnm{Brunton}, \binits{S.L.}},
\oauthor{\bsnm{Chen}, \binits{Y.}},
\oauthor{\bsnm{How}, \binits{J.P.}}:
Robust, High-Rate Trajectory Tracking on Insect-Scale Soft-Actuated Aerial Robots with Deep-Learned Tube MPC
(2022)
\end{botherref}
\endbibitem

\bibitem[\protect\citeauthoryear{Dhingra et~al.}{2020}]{ddhingraTrimming}
\begin{barticle}
\bauthor{\bsnm{Dhingra}, \binits{D.}},
\bauthor{\bsnm{Chukewad}, \binits{Y.M.}},
\bauthor{\bsnm{Fuller}, \binits{S.B.}}:
\batitle{A device for rapid, automated trimming of insect-sized flying robots}.
\bjtitle{IEEE Robotics and Automation Letters}
\bvolume{5}(\bissue{2}),
\bfpage{1373}--\blpage{1380}
(\byear{2020})
\doiurl{10.1109/LRA.2020.2967318}
\end{barticle}
\endbibitem

\bibitem[\protect\citeauthoryear{Bena et~al.}{2023}]{Bee++2023}
\begin{barticle}
\bauthor{\bsnm{Bena}, \binits{R.M.}},
\bauthor{\bsnm{Yang}, \binits{X.}},
\bauthor{\bsnm{Calderón}, \binits{A.A.}},
\bauthor{\bsnm{Pérez-Arancibia}, \binits{N.O.}}:
\batitle{High-performance six-{DOF} flight control of the {Bee}$^{++}$: An inclined-stroke-plane approach}.
\bjtitle{IEEE Transactions on Robotics}
\bvolume{39}(\bissue{2}),
\bfpage{1668}--\blpage{1684}
(\byear{2023})
\doiurl{10.1109/TRO.2022.3218260}
\end{barticle}
\endbibitem

\bibitem[\protect\citeauthoryear{Chukewad et~al.}{2018}]{chukewad2018new}
\begin{bchapter}
\bauthor{\bsnm{Chukewad}, \binits{Y.M.}},
\bauthor{\bsnm{Singh}, \binits{A.T.}},
\bauthor{\bsnm{James}, \binits{J.M.}},
\bauthor{\bsnm{Fuller}, \binits{S.B.}}:
\bctitle{A new robot fly design that is easier to fabricate and capable of flight and ground locomotion}.
In: \bbtitle{2018 IEEE/RSJ International Conference on Intelligent Robots and Systems (IROS)},
pp. \bfpage{4875}--\blpage{4882}
(\byear{2018}).
\bcomment{IEEE}
\end{bchapter}
\endbibitem

\bibitem[\protect\citeauthoryear{Bouabdallah et~al.}{2004}]{quad_LQR_2004}
\begin{bchapter}
\bauthor{\bsnm{Bouabdallah}, \binits{S.}},
\bauthor{\bsnm{Noth}, \binits{A.}},
\bauthor{\bsnm{Siegwart}, \binits{R.}}:
\bctitle{{PID} vs {LQ} control techniques applied to an indoor micro quadrotor}.
In: \bbtitle{2004 IEEE/RSJ International Conference on Intelligent Robots and Systems (IROS) (IEEE Cat. No.04CH37566)},
vol. \bseriesno{3},
pp. \bfpage{2451}--\blpage{24563}
(\byear{2004}).
\doiurl{10.1109/IROS.2004.1389776}
\end{bchapter}
\endbibitem

\bibitem[\protect\citeauthoryear{Jafferis et~al.}{2015}]{Prestack_Jafferis_2015}
\begin{botherref}
\oauthor{\bsnm{Jafferis}, \binits{N.T.}},
\oauthor{\bsnm{Smith}, \binits{M.J.}},
\oauthor{\bsnm{Wood}, \binits{R.J.}}:
Design and manufacturing rules for maximizing the performance of polycrystalline piezoelectric bending actuators
\textbf{24}(6),
065023
(2015)
\doiurl{10.1088/0964-1726/24/6/065023}
\end{botherref}
\endbibitem

\bibitem[\protect\citeauthoryear{Chukewad et~al.}{2021}]{ChukewadTRO}
\begin{barticle}
\bauthor{\bsnm{Chukewad}, \binits{Y.M.}},
\bauthor{\bsnm{James}, \binits{J.}},
\bauthor{\bsnm{Singh}, \binits{A.}},
\bauthor{\bsnm{Fuller}, \binits{S.}}:
\batitle{Robofly: An insect-sized robot with simplified fabrication that is capable of flight, ground, and water surface locomotion}.
\bjtitle{IEEE Transactions on Robotics}
\bvolume{37}(\bissue{6}),
\bfpage{2025}--\blpage{2040}
(\byear{2021})
\doiurl{10.1109/TRO.2021.3075374}
\end{barticle}
\endbibitem

\bibitem[\protect\citeauthoryear{Alavilli et~al.}{2024}]{alavilli2024tinympc}
\begin{bchapter}
\bauthor{\bsnm{Alavilli}, \binits{A.}},
\bauthor{\bsnm{Nguyen}, \binits{K.}},
\bauthor{\bsnm{Schoedel}, \binits{S.}},
\bauthor{\bsnm{Plancher}, \binits{B.}},
\bauthor{\bsnm{Manchester}, \binits{Z.}}:
\bctitle{Tinympc: Model-predictive control on resource-constrained microcontrollers}.
In: \bbtitle{IEEE International Conference on Robotics and Automation (ICRA)},
\bconflocation{Yokohama, Japan}
(\byear{2024})
\end{bchapter}
\endbibitem

\bibitem[\protect\citeauthoryear{Englert et~al.}{2018}]{englert2018software}
\begin{botherref}
\oauthor{\bsnm{Englert}, \binits{T.}},
\oauthor{\bsnm{Völz}, \binits{A.}},
\oauthor{\bsnm{Mesmer}, \binits{F.}},
\oauthor{\bsnm{Rhein}, \binits{S.}},
\oauthor{\bsnm{Graichen}, \binits{K.}}:
A software framework for embedded nonlinear model predictive control using a gradient-based augmented Lagrangian approach (GRAMPC)
(2018)
\end{botherref}
\endbibitem

\bibitem[\protect\citeauthoryear{Ma et~al.}{2013}]{ma2013controlled}
\begin{barticle}
\bauthor{\bsnm{Ma}, \binits{K.Y.}},
\bauthor{\bsnm{Chirarattananon}, \binits{P.}},
\bauthor{\bsnm{Fuller}, \binits{S.B.}},
\bauthor{\bsnm{Wood}, \binits{R.J.}}:
\batitle{Controlled flight of a biologically inspired, insect-scale robot}.
\bjtitle{Science}
\bvolume{340}(\bissue{6132}),
\bfpage{603}--\blpage{607}
(\byear{2013})
\end{barticle}
\endbibitem

\bibitem[\protect\citeauthoryear{Yang et~al.}{2019}]{beeplus}
\begin{barticle}
\bauthor{\bsnm{Yang}, \binits{X.}},
\bauthor{\bsnm{Chen}, \binits{Y.}},
\bauthor{\bsnm{Chang}, \binits{L.}},
\bauthor{\bsnm{Calderón}, \binits{A.A.}},
\bauthor{\bsnm{Pérez-Arancibia}, \binits{N.O.}}:
\batitle{Bee+: A 95-mg four-winged insect-scale flying robot driven by twinned unimorph actuators}.
\bjtitle{IEEE Robotics and Automation Letters}
\bvolume{4}(\bissue{4}),
\bfpage{4270}--\blpage{4277}
(\byear{2019})
\doiurl{10.1109/LRA.2019.2931177}
\end{barticle}
\endbibitem

\bibitem[\protect\citeauthoryear{Weber et~al.}{2024}]{AaronMapping2024}
\begin{botherref}
\oauthor{\bsnm{Weber}, \binits{A.}},
\oauthor{\bsnm{Dhingra}, \binits{D.}},
\oauthor{\bsnm{Fuller}, \binits{S.B.}}:
A flexured-gimbal 3-axis force-torque sensor reveals minimal cross-axis coupling in an insect-sized flapping-wing robot
(2024)
\end{botherref}
\endbibitem

\bibitem[\protect\citeauthoryear{Malka et~al.}{2014}]{malka2014}
\begin{bchapter}
\bauthor{\bsnm{Malka}, \binits{R.}},
\bauthor{\bsnm{Desbiens}, \binits{A.L.}},
\bauthor{\bsnm{Chen}, \binits{Y.}},
\bauthor{\bsnm{Wood}, \binits{R.J.}}:
\bctitle{Principles of microscale flexure hinge design for enhanced endurance}.
In: \bbtitle{2014 IEEE/RSJ International Conference on Intelligent Robots and Systems},
pp. \bfpage{2879}--\blpage{2885}
(\byear{2014}).
\doiurl{10.1109/IROS.2014.6942958}
\end{bchapter}
\endbibitem

\bibitem[\protect\citeauthoryear{Fuller}{2019}]{fuller2019four}
\begin{barticle}
\bauthor{\bsnm{Fuller}, \binits{S.B.}}:
\batitle{Four wings: An insect-sized aerial robot with steering ability and payload capacity for autonomy}.
\bjtitle{IEEE Robotics and Automation Letters}
\bvolume{4}(\bissue{2}),
\bfpage{570}--\blpage{577}
(\byear{2019})
\end{barticle}
\endbibitem

\bibitem[\protect\citeauthoryear{Anderson and Moore}{2007}]{andersonoptimal2007}
\begin{botherref}
\oauthor{\bsnm{Anderson}, \binits{B.D.O.}},
\oauthor{\bsnm{Moore}, \binits{J.B.}}:
Optimal control: Linear quadratic methods
(2007)
\end{botherref}
\endbibitem

\end{thebibliography}

\end{document}